\documentclass{article}
\usepackage{amssymb}
\usepackage{spconf,amsmath,graphicx}
\usepackage{enumitem}
\usepackage{booktabs}
\usepackage{multirow}

\DeclareMathAlphabet\mathbfcal{OMS}{cmsy}{b}{n}

\title{SPEECH-DRIVEN FACIAL ANIMATION USING POLYNOMIAL FUSION OF FEATURES}
\name{
\begin{tabular}{@{}c@{}}
Triantafyllos Kefalas$^{\star}$ \qquad 
Konstantinos Vougioukas$^{\star}$ \qquad 
Yannis Panagakis$^{\dagger}$\qquad 
Stavros Petridis$^{\star \ddagger}$ \\ 
Jean Kossaifi$^{\star \ddagger}$ \qquad 
Maja Pantic$^{\star \ddagger}$
\end{tabular}
}

\address{$^{\star}$ Department of Computing, Imperial College London, UK \\
$^{\dagger}$ Department of Informatics and Telecommunications, University of Athens, Greece \\
$^{\ddagger}$ Samsung AI Centre, Cambridge, UK}

\begin{document}
%
\maketitle
\begin{abstract}
Speech-driven facial animation involves using a speech signal to generate realistic videos of talking faces. Recent deep learning approaches to facial synthesis rely on extracting low-dimensional representations and concatenating them, followed by a decoding step of the concatenated vector. This accounts for only first-order interactions of the features and ignores higher-order interactions. In this paper we propose a polynomial fusion layer that models the joint representation of the encodings by a higher-order polynomial, with the parameters modelled by a tensor decomposition. We demonstrate the suitability of this approach through experiments on generated videos evaluated on a range of metrics on video quality, audiovisual synchronisation and generation of blinks.
\end{abstract}
\section{Introduction}
\label{sec:intro}

The problem of speech-driven facial animation entails generating realistic videos of talking faces based on a speech input \cite{end2endSpeechDrivenFacialAnimation, sda_ijcv, you_said_that, mocogan, expressive_sda}. Facial animation is an important part of computer generated imagery (CGI) applications, such as video games and virtual assistants \cite{expressive_sda, sda_ijcv}. Considered one of the most challenging problems in computer graphics, facial animation involves the coordinated movement of hundreds of muscles \cite{expressive_sda} and expresses a wide range of emotions, in addition to the underlying speech. Furthermore, generating realistic videos requires synchronisation of lip movements with speech as well as natural facial expressions such as blinks \cite{sda_ijcv}.

CGI traditionally employs face capture methods for facial synthesis \cite{sda_ijcv}, which are costly and require manual labor. To alleviate this, recent research has focused on automatic generation of video with machine learning \cite{end2endSpeechDrivenFacialAnimation, sda_ijcv, you_said_that, mocogan, expressive_sda}. MoCoGAN \cite{mocogan} models motion and content as separate latent spaces, which are learned in an unsupervised way using both image and video discriminators. The Speech2Vid model \cite{you_said_that} generates a video of a target face which is lip synced with the audio signal. This model uses CNNs on Mel-Frequency Cepstral Coefficients (MFCCs) of the audio input as well as a frame deblurring network to preserve high-frequency visual content. Vougioukas et al. \cite{sda_ijcv} propose a deep architecture for speech-driven facial synthesis that takes into account facial expression dynamics in the course of speech. This is achieved by including a temporal noise input signal as well as by using multiple discriminators to control for frame quality, audio-visual synchronisation and overall video quality.

The aforementioned methods extract low-dimensional representations from the inputs and combine them with a simple concatenation operation. Concatenating multiple inputs to produce a joint representation has several limitations, such as accounting for only the first-order interactions among inputs and excluding higher-order interactions. In addition, the statistics of the concatenated vector can be dominated by a subset of the representations following any arbitrary scaling of the latter. Furthermore, we often wish to capture invariances in the data to perform a given task and a concatenated joint representation may not reflect this. For example, in speaker identification applications it has been observed that concatenating audio and visual features makes the joint representation sensitive to large facial movements \cite{attention_guided_deep_av_fusion}.

In this work, we abstract away the concatenation operation into a new \textit{polynomial fusion layer} that models the interactions between the low-dimensional representations of the multiple views (visual and audio). We propose modelling the joint representation as a higher-order polynomial of the inputs, whose parameters are naturally represented by tensors. To alleviate the exponential space complexity of tensors, we model the parameters of the polynomial using tensor factorizations \cite{kolda2009}. Previous work on polynomial fusion of features \cite{polygan} modelled within-feature interactions from a single view, whereas our method focuses on between-feature interactions from multiple views. To our knowledge, this is the first work in the literature that uses tensor factorizations and concepts from multi-view learning \cite{multi_view_machines} \cite{multilinear_multitask_learning} to compute a joint representation for a \textit{generative} task.

\subsection{Notation and preliminaries}

The notation of \cite{kolda2009} is adopted in this paper. The sets of integers and real numbers are denoted by $\mathbb{Z}$ and $\mathbb{R}$ respectively. Matrices and vectors are denoted by bold uppercase (e.g. $\mathbf{X} \in \mathbb{R}^{{I_1} \times {I_2}}$) and lower case letters (e.g. $\mathbf{x} \in \mathbb{R}^{I_1}$) respectively. Tensors are higher-dimensional generalisations of vectors (first-order tensors) and matrices (second-order tensors) denoted by bold calligraphic capital letters (e.g. $\mathbfcal{X}$). The order of a tensor is the number of indices required to address its elements. A real-valued tensor $\mathbfcal{X} \in \mathbb{R}^{I_1 \times I_2 \times ... \times I_N}$ of order $N$ is defined over the tensor product of $N$ vector spaces, where $I_n \in \mathbb{Z}$ for $n = 1, 2, ... N$.

A set of $N$ real matrices (vectors) of varying dimensionality is denoted by $\{ \mathbf{X}^{(n)} \in \mathbb{R}^{I_n \times J_n} \}_{n=1}^N$\big{(} $\{ \mathbf{x}^{(n)} \in \mathbb{R}^{I_n} \}_{n=1}^N$\big{)}. An order-$N$ tensor has $rank$-$1$ when it is expressed as the outer product of $N$ vectors, i.e. $\mathbfcal{X} = \mathbf{x}^{(1)} \circ \mathbf{x}^{(2)} \circ ... \circ \mathbf{x}^{(N)}$ where $\{ \mathbf{x}^{(n)} \in \mathbb{R}^{I_n} \}_{n=1}^N$.

The mode-$n$ unfolding matrix of tensor $\mathbfcal{X} \in \mathbb{R}^{I_1 \times I_2 \times ... \times I_N}$ reorders the elements of the tensor into a matrix $\mathbf{X}_{(n)} \in \mathbb{R}^{I_n \times \bar{I}_n}$ where $\bar{I}_n = \prod_{{\small{\substack{k=1\\k \neq n}}}}^N I_k$. The mode-$n$ vector product of a tensor $\mathbfcal{X} \in \mathbb{R}^{I_1 \times I_2 \times ... \times I_N}$ and a vector $\mathbf{x} \in \mathbb{R}^{I_n}$ results to an order $N-1$ tensor and is denoted by $\mathbfcal{X} \times_n \mathbf{x} \in \mathbb{R}^{I_1 \times I_2 \times ... \times I_{n-1} \times I_{n+1} \times ... \times I_N}$.
The Kronecker product and Khatri-Rao product, defined in \cite{kolda2009} are denoted by $\otimes$ and $\odot$ respectively. Finally, the Frobenius norm is denoted by $|| \cdot ||_F$. The reader is referred to \cite{kolda2009} for a detailed survey of tensors and tensor notation.

\section{Methodology}

\subsection{Pipeline overview}

\begin{figure}[ht]
\includegraphics[width=8.5cm]{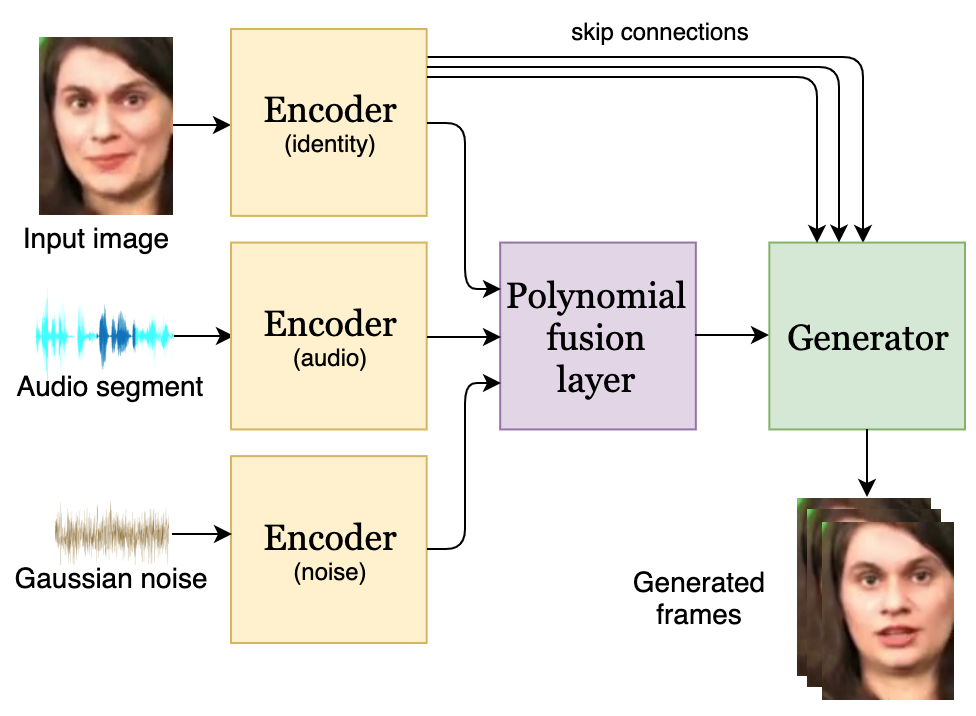}
\caption{Pipeline overview}
\label{pipeline_overview}
\end{figure}

The pipeline follows an encoder-decoder structure, inspired by \cite{sda_ijcv}, and is illustrated in Fig. \ref{pipeline_overview}. The three temporal encoders extract representations of the speaker identity, the audio segment and spontaneous facial expressions. The polynomial fusion layer combines the three encodings which are then fed into the generator produce the frames.

The identity encoder takes as input a single face image and passes it through a 6-layer CNN, where each layer has strided 2D convolutions, batch normalization and ReLU activation functions. The audio encoder receives a 0.2s audio signal and passes it through a network of 1D convolutions with batch normalization and ReLU activation functions, followed by a 1-layer GRU. The noise encoder receives a gaussian noise input which is passed through a 1-layer GRU. The purpose of the noise encoder is to introduce some natural variability in the face generation such as blinks and eyebrow movements. 

The polynomial fusion layer generates a joint representation of the three encodings which is then fed into the U-Net based \cite{unet_paper} generator, along with skip connections from the intermediate layers of the identity encoder, to produce a frame from a video sequence.

Furthermore, we employ three CNN-based disciminators during training to control for frame quality, plausibility of the video (frame sequence) and audiovisual synchronisation.

\subsection{Polynomial fusion layer}
Let $\mathbf{z}_a \in \mathbb{R}^{a}$, $\mathbf{z}_d \in \mathbb{R}^{d}$ and $\mathbf{z}_n \in \mathbb{R}^{n}$ be the encoder outputs (encodings) for the audio, identity and noise respectively. The polynomial fusion layer is a function that takes the audio and identity encodings as input and returns a vector $\mathbf{\Tilde{z}} \in \mathbb{R}^m$, which is the joint representation:

$$ \mathbf{\Tilde{z}} = f(\mathbf{z}_a, \mathbf{z}_d)$$
where $m$ is a hyperparameter.

We propose modelling the interactions of the audio and identity encodings using a higher-order polynomial:

\begin{equation} \label{eq:polynomial}
\mathbf{\Tilde{z}} = \mathbf{b} + \mathbf{W}^{[a]}\mathbf{z}_a + \mathbf{W}^{[d]}\mathbf{z}_d + \mathbfcal{W}^{[a, d]} \times_2\mathbf{z}_a \times_3\mathbf{z}_d 
\end{equation}
where $\mathbf{b}\in \mathbb{R}^m$ is the global bias, $\mathbf{W}^{[a]} \in \mathbb{R}^{m \times a}$, $\mathbf{W}^{[d]} \in \mathbb{R}^{m \times d}$ are the first-order interaction matrices for audio and identity respectively and $\mathbfcal{W}^{[a, d]} \in \mathbb{R}^{m \times a \times d}$ is the tensor of second-order interactions between audio and identity. Finally, we concatenate the noise embedding to the joint representation to obtain the input for the generator:

\begin{equation} \label{eq:1}
\mathbf{z} = [\mathbf{\Tilde{z}}, \mathbf{z}_n]
\end{equation}
where $\mathbf{z} \in \mathbb{R}^{c}$, $c = m + n$.

However, the number of parameters in (\ref{eq:polynomial}) is exponential with respect to the number of input embeddings. To reduce the complexity, it is common practice to assume a low-rank structure of the parameters which can be captured using tensor factorizations \cite{low_rank_tensor_networks}. We propose several ways to do this, as follows:
\newline
\newline
\textbf{Coupled matrix-tensor factorizations}: In this approach we model the dependence between the first-order and second-order interactions by the sharing of column and/or row spaces. By allowing the first-order and second-order interactions to share their column space (column mode) we can factorize (\ref{eq:polynomial}) as follows:

\begin{equation} \label{eq:3}
\mathbf{W}^{[a]} = \mathbf{U}\mathbf{V}^{[a] T}
\end{equation}
\begin{equation} \label{eq:4}
\mathbf{W}^{[d]} = \mathbf{U}\mathbf{V}^{[d] T}
\end{equation}
where $\mathbf{U} \in \mathbb{R}^{m \times k}$ describes the shared column space, $\mathbf{V}^{[a]} \in \mathbb{R}^{a \times k}$, $\mathbf{V}^{[d]} \in \mathbb{R}^{d \times k}$ describe the audio-specific and identity specific row spaces respectively. 

Furthermore, we perform the Canonical Polyadic (CP) \cite{cp_decomposition} decomposition on $\mathbfcal{W}^{[a, d]}$ defined as:

\begin{equation} \label{eq:5}
\mathbf{W}^{[a, d]}_{(1)} = \mathbf{B}^{(1)}(\mathbf{B}^{(3)}\odot \mathbf{B}^{(2)})^T
\end{equation}
where $\mathbf{B}^{(1)} \in \mathbb{R}^{m \times k}$, $\mathbf{B}^{(2)} \in \mathbb{R}^{a \times k}$, $\mathbf{B}^{(3)} \in \mathbb{R}^{d \times k}$ are the factor matrices. We also apply the constraint $\mathbf{B}^{(1)} := \mathbf{U} $ to enable the sharing of the column space.

We can perform further parameter sharing by assuming that the mode-2 and mode-3 factor matrices are equivalent to the matrices describing the row spaces, i.e.: $\mathbf{B}^{(2)} := \mathbf{V}^{[a]}$, $\mathbf{B}^{(3)} := \mathbf{V}^{[d]}$ (6b)
\newline
\newline
\textbf{Joint factorization of interactions}: In this approach, we collect the parameters in (\ref{eq:polynomial}) into a single parameter tensor to perform joint factorization of all-order interactions. Let $\mathbfcal{W} \in \mathbb{R}^{m \times (a+1) \times (d+1)}$ be the single parameter tensor defined as follows: $\mathbfcal{W}_{:, a+1, d+1} := \mathbf{b}$, $
\mathbfcal{W}_{:, 1:a, d+1} := \mathbf{W}^{[a]}$, $ \mathbfcal{W}_{:, a+1, 1:d} := \mathbf{W}^{[d]}$, $\mathbfcal{W}_{:, 1:a, 1:d} := \mathbfcal{W}^{[a, d]}$, where $:$ indicates all elements of a mode of the tensor and $1:x$ indicates all elements from the first to the $x$th element of that mode.

Using $\mathbfcal{W}$ we then compute the joint representation:

\begin{equation} \label{eq:7}
\mathbf{\Tilde{z}} = \mathbfcal{W} \times_2 \mathbf{\phi}(\mathbf{z}_a) \times_3 \mathbf{\phi}(\mathbf{z}_d)
\end{equation}
where $\phi(\mathbf{x}) := [\mathbf{x}, 1]$ concatenates a $1$ to the end of the input vector. The last step is required for the computation of interactions of all orders as per \cite{multi_view_machines}.

We investigate the Canonical Polyadic (CP) \cite{cp_decomposition} and Tucker decompositions \cite{tucker_decomposition} respectively. The CP decomposition for this model is defined as follows:

\begin{equation} \label{eq:8}
\mathbf{W}_{(1)} = \mathbf{A}^{(1)}(\mathbf{A}^{(3)}\odot \mathbf{A}^{(2)})^T
\end{equation}
where $\{\mathbf{A}^{(n)}\}_{n=1}^3$ are the factor matrices where $\mathbf{A}^{(1)} \in \mathbb{R}^{m \times k}$, $\mathbf{A}^{(2)} \in \mathbb{R}^{(a+1) \times k}$, $\mathbf{A}^{(3)} \in \mathbb{R}^{(d+1) \times k}$, and $k$ is the tensor rank.

The Tucker decomposition can be expressed in tensor notation as follows:

\begin{equation} \label{eq:9}
\mathcal{W} = \mathcal{G} \times_1 \mathbf{U}^{(1)} \times_2 \mathbf{U}^{(2)} \times_3 \mathbf{U}^{(3)}
\end{equation}
where $\mathcal{G} \in \mathbb{R}^{k_1 \times k_2 \times k_3}$ is the Tucker core, $\{\mathbf{U}^{(n)}\}_{n=1}^3$ are the factor matrices where
$\mathbf{U}^{(1)} \in \mathbb{R}^{m \times k_1}$, $\mathbf{U}^{(2)} \in \mathbb{R}^{(a+1) \times k_2}$, $\mathbf{U}^{(3)} \in \mathbb{R}^{(d+1) \times k_3}$, and $(k_1, k_2, k_3)$ is the multilinear rank.

\subsection{Objective function}

Let $G$ denote the parameters of the polynomial fusion layer and the generator. We assume that the encoders are pre-trained and fixed. Let $D = \{ D_{frame}, D_{seq}, D_{sync} \}$ denote the parameters of the three discriminators (frame, video sequence, synchronisation). The adversarial losses for each discriminator is are the following:

\footnotesize
\begin{multline} \label{eq:10}
\mathcal{L}_{adv}^{frame} = \mathbb{E}_{x \sim P_d}[\log D_{frame}\big{(}S(x), x_1\big{)}] + \\
\mathbb{E}_{z \sim P_z}[\log (1 - D_{frame}\big{(}S(G(z)), x_1\big{)}]
\end{multline}
\begin{multline} \label{eq:11}
\mathcal{L}_{adv}^{seq} = \mathbb{E}_{x \sim P_d}[\log D_{seq}\big{(}x, a\big{)}] + \\
\mathbb{E}_{z \sim P_z}[\log (1 - D_{seq}\big{(}S(G(z)), a\big{)}]
\end{multline}
\begin{multline} \label{eq:12}
\mathcal{L}_{adv}^{sync} = \mathbb{E}_{x \sim P_d}[\log D_{sync}\big{(}p_{in}\big{)}] + \frac{1}{2}\mathbb{E}_{x \sim P_d} \big{[}\log1 - D_{sync}(p_{out})\big{]} \\ 
+ \frac{1}{2}\mathbb{E}_{z \sim P_z}[\log (1 - D_{sync}\big{(}S_{snip}(p_f)\big{)}]
\end{multline}
\normalsize
where $x$ is a sample video, $S(x)$ is a sampling function, $a$ is the corresponding audio, $\{ p_{in}, p_{out} \}$ are in and out of sync pairs of the ground truth video and $p_f$ is a generated video and its corresponding audio. We also assign a corresponding weight to each of the above adversarial losses to obtain the overall adversarial loss: \\
$\mathcal{L}_{adv}:= 
\lambda_{frame}\mathcal{L}_{adv}^{frame} + \lambda_{seq}\mathcal{L}_{adv}^{seq} + \lambda_{sync}\mathcal{L}_{adv}^{sync}$.

Furthermore, we include an $l_1$-norm reconstruction loss, $\mathcal{L}_{L_1}$ on the bottom half of the frame to encourage the correct mouth movements. Finally, we add a Frobenius norm penality term, $\Omega$ applied to the parameters of the fusion layer.

The final objective function is then the following:

\begin{equation} \label{eq:13}
\min_G \max_D \mathcal{L}_{adv} + \lambda_1 \mathcal{L}_{L_1} + \lambda_2\Omega 
\end{equation}
where $\lambda_1, \lambda_2, \lambda_{frame}, \lambda_{seq}$ and $\lambda_{sync}$ are hyperparameters.

\section{Experimental evaluation}
\label{sec:pagestyle}

We conducted experiments on the GRID \cite{grid_database}, CREMA-D \cite{crema_databse} and TCD TIMIT \cite{tcd_timit_databse} datasets.
GRID contains 33 speakers uttering 1000 short structured sentences. CREMA-D contains data from 91 actors uttering 12 sentences acted out for different emotions. TCD TIMIT contains data from 59 speakers uttering a selection of approximately 100 sentences from the TIMIT \cite{timit_database} dataset.

\begin{table*}[ht]
\label{results_table}
\resizebox{\textwidth}{!}{%
\begin{tabular}{@{}clcccccccc@{}}
\toprule
\multirow{2}{*}{Dataset} & \multicolumn{1}{c}{\multirow{2}{*}{Method}} & \multicolumn{8}{c}{Metrics} \\ \cmidrule(l){3-10} 
 & \multicolumn{1}{c}{} & PSNR & SSIM & CPBD & ACD & AV Offset & AV confidence & blinks/sec & blink duration (sec.) \\ \midrule
\multirow{6}{*}{GRID} & PF-Tucker & 27.360 & 0.831 & 0.260 & 1.11 x $10^{-4}$  & 1 & 7.1 & 0.37 & 0.38 \\
 & PF-CP & \textbf{27.619} & \textbf{0.863} & 0.259 & 1.18 x $10^{-4}$  & 1 & 7.3 & 0.32 & 0.36 \\
 & PF-CMF & 27.529 & 0.836 & \textbf{0.268} & \textbf{8.63 x} $\textbf{10}^\textbf{{-5}}$  & 1 & \textbf{7.4} & 0.33 & 0.36 \\
 & PF-CMF-SR & 27.608 & 0.833 & 0.253 & 9.74 x $10^{-5}$ & 1 & 7.3 & 0.37 & 0.32 \\
 & Vougioukas et al, 2019 & 27.100 & 0.818 & \textbf{0.268} & 1.47 x $10^{-4}$ & 1 & \textbf{7.4} & 0.45 & 0.36 \\
 & Speech2Vid & 22.662 & 0.720 & 0.255 & 1.48 x $10^{-4}$ & 1 & 5.3 & 0.00 & 0.00 \\ \cmidrule(l){1-10}
\multirow{6}{*}{TCD-TIMIT} & PF-Tucker & 23.958 & 0.722 & 0.283 & 2.69 x $10^{-4}$ & 1 & 5.5 & 0.13 & 0.31 \\
 & PF-CP & \textbf{24.403} & \textbf{0.737} & 0.304 & 2.04 x $10^{-4}$ & 1 & \textbf{5.6} & 0.11 & 0.28 \\
 & PF-CMF & 24.095 & 0.723 & 0.286 & 2.65 x $10^{-4}$ & 1 & 5.5 & 0.10 & 0.30 \\
 & PF-CMF-SR & 23.871 & 0.719 & 0.285 & 2.94 x $10^{-4}$ & 1 & 5.2 & 0.07 & 0.31 \\
 & Vougioukas et al, 2019 & 24.243 & 0.730 & \textbf{0.308} & \textbf{1.76 x} $\textbf{10}^\textbf{-4}$ & 1 & \textbf{5.5} & 0.19 & 0.33 \\
 & Speech2Vid & 20.305 & 0.658 & 0.211 & 1.81 x $10^{-4}$ & 1 & 4.6 & 0.00 & 0.00 \\ \cmidrule(l){1-10}
\multirow{6}{*}{CREMA-D} & PF-Tucker & 23.272 & 0.691 & 0.221 & 1.96 x $10^{-4}$ & 2 & 5.1 & 0.25 & 0.39 \\
 & PF-CP & 23.502 & 0.698 & \textbf{0.233} & 1.69 x $10^{-4}$ & 2 & 5.1 & 0.17 & 0.37 \\
 & PF-CMF & 23.277 & 0.689 & 0.208 & 1.73 x $10^{-4}$ & 2 & 5.1 & 0.38 & 0.35 \\
 & PF-CMF-SR & 23.457 & 0.697 & 0.214 & 1.63 x $10^{-4}$ & 2 & 5.3 & 0.18 & 0.32 \\
 & Vougioukas et al, 2019 & \textbf{23.565} & \textbf{0.700} & 0.216 & \textbf{1.40 x }$\textbf{10}^\textbf{-4}$ & 2 & \textbf{5.5} & 0.25 & 0.26 \\
 & Speech2Vid & 22.190 & \textbf{0.700} & 0.217 & 1.73 x $10^{-4}$ & 1 & 4.7 & 0.00 & 0.00 \\ \bottomrule
\end{tabular}
}
\caption{Evaluation summary of the proposed method in comparison with Vougioukas et al., 2019 and Speech2Vid}
\end{table*}

We use a 50\%-20\%-30\% training, validation and test set split on the speakers of GRID, and a 70\%-15\%-15\% datset split for CREMA-D. For TCD TIMIT we use the recommended split, excluding some test set speakers to be used as validation.

\subsection{Training protocol}
We ran experiments on the above datasets using the following methodologies for the polynomial fusion layer:
\begin{itemize}[noitemsep, nolistsep]
\item Tucker decomposition as per (\ref{eq:9}) (PF-Tucker)
\item CP decomposition as per (\ref{eq:8}) (PF-CP)
\item Coupled matrix-tensor factorization as per (\ref{eq:3}), (\ref{eq:4}), (\ref{eq:5}) (PF-CMF)
\item Coupled matrix-tensor factorization with sharing of row space matrices as per (\ref{eq:3}), (\ref{eq:4}), (\ref{eq:5}), (6b) (PF-CMF-SR)
\end{itemize}

We set $a = 256, d = 128, n = 10$ and trained on video sequences of 3 seconds with frame size $128 \times 96$ as per \cite{sda_ijcv}. For all models, $c = a + d + n = 394$, implying $m = 384$, given that $c = m + n$ by construction. For the Tucker decomposition above, we set $(k_1, k_2, k_3) = (192, 128, 64)$, corresponding to half the dimensionality along each mode. We set $k = d = 128$ for all remaining methodologies.

For the three encoders, we used the pre-trained weights from the experiments of Vougioukas et al. \cite{sda_ijcv} and these remained fixed during training. We performed Gaussian initialization on the parameters of the polynomial fusion layer, the generator and the discriminators. We set the hyperparameters $\lambda_1 = 600, \lambda_{frame} = 1, \lambda_{seq}=0.2, \lambda_{sync}=0.8$ as in \cite{sda_ijcv} and $\lambda_2 = 100$. The models were trained with Adam \cite{adam_paper} using learning rates of $1 \times 10^{-4}$ for $G$ and $D_{frame}$ and $1 \times 10^{-5}$ for $D_{seq}$ and $D_{sync}$. Training was carried out until, by inspection, there was no noticeable improvement in the video quality on the validation set for 5 epochs.

We also compare our models with Vougioukas et al. \cite{sda_ijcv} which uses concatenation as its fusion layer, and the Speech2Vid model \cite{you_said_that}, which is trained on the Voxceleb \cite{voxceleb_paper} and LRW \cite{lip_reading_in_the_wild} datasets.

We evaluate the reconstruction quality of the videos the results using the peak signal to noise ratio (PSNR) and structural similarity index (SSIM). In addition, we measure the sharpness of the frames using the cumulative probability blur detection measure \cite{cpbd_paper}. We perform face verification on the videos using the average content distance (ACD) metric using OpenFace \cite{openface}. For PSNR and SSIM, higher values indicate better quality whereas for ACD a lower value indicates a better capturing of the identity. Furthermore, we evaluate the audiovisual synchronisation of the videos using SyncNet \cite{syncnet_paper}, where lower AV offset (measured in frames) indicates better synchronisation and higher AV confidence indicates stronger audiovisual correlation. For reference we also evaluate the blinks in generated videos using the blink detector in \cite{sda_ijcv}. The results are shown on Table \ref{results_table}.

From the experiments above, we observe that our model outperforms the Speech2Vid model and is comparable to the state-of-the-art of Vougioukas et al, 2019. In particular, our model is able to generate natural blinks at rates comparable to real-life values (estimated at 0.28 blinks/sec \cite{blink_study}).

\section{Conclusion}
\label{sec:majhead}

In this paper we present a polynomial fusion layer that produces a joint low-dimensional representation of encodings of audio and visual identity using a higher-order polynomial. The parameters of the polynomial are modelled by a low-rank tensor decomposition. We have investigated this method for the task of speech-driven facial animation, where the joint representation is fed into a generator to produce a sequence of talking faces. We demonstrate the comparability of the proposed method with the state-of-the art on several metrics from experiments on three audiovisual datasets.

\clearpage

\vfill\pagebreak
\clearpage

\newpage
\bibliographystyle{IEEEbib}
\bibliography{strings,refs}

\end{document}